\pdfoutput=1

\documentclass[11pt]{article}

\usepackage{acl}

\usepackage{times}
\usepackage{latexsym}
\usepackage{graphicx}
\usepackage{xcolor}
\usepackage{soul}
\usepackage{todonotes}
\usepackage{xspace}

\newcommand{\ds}{WikiBio\xspace}

\usepackage[T1]{fontenc}

\usepackage[utf8]{inputenc}

\usepackage{microtype}
\usepackage{booktabs}

\title{\ds: a Semantic Resource\\ for the Intersectional Analysis of Biographical Events}

\author{Marco Antonio Stranisci\textsuperscript{\includegraphics[height=1.0em]{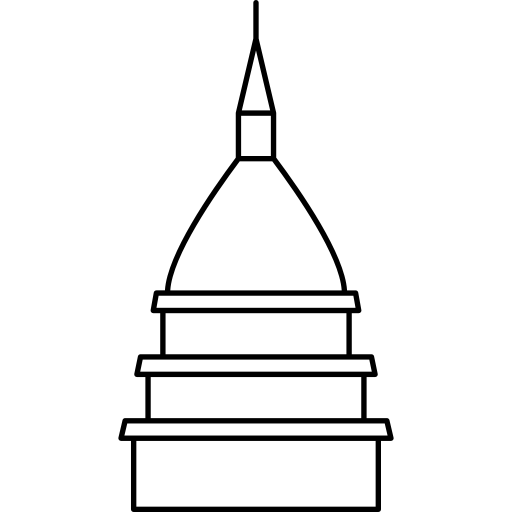}}\textsuperscript{\includegraphics[height=0.5em]{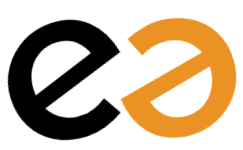}}, Rossana Damiano\textsuperscript{\includegraphics[height=1.0em]{figures/loghi/unito.png}}, Enrico Mensa\textsuperscript{\includegraphics[height=1.0em]{figures/loghi/unito.png}}\\ \textbf{Viviana  Patti\textsuperscript{\includegraphics[height=1.0em]{figures/loghi/unito.png}}, Daniele Paolo Radicioni\textsuperscript{\includegraphics[height=1.0em]{figures/loghi/unito.png}}, Tommaso Caselli\textsuperscript{\includegraphics[height=1.0em]{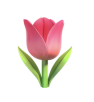}}} \\
\textsuperscript{\includegraphics[height=1.0em]{figures/loghi/unito.png}}Dipartimento di Informatica, Università degli Studi di Torino, Italy \\
\textsuperscript{\includegraphics[height=1.0em]{figures/loghi/tom.png}}CLCG, University of  Groningen \\
\textsuperscript{\includegraphics[height=0.5em]{figures/loghi/favicon.png}}aequa-tech, Turin, Italy \\
{\tt marcoantonio.stranisci@unito.it}
} 

\begin{document}
\maketitle
\begin{abstract}
Biographical event detection is a relevant task for the exploration and comparison of the ways in which people's lives are told and represented. In this sense, it may support several applications in digital humanities and in works aimed at exploring bias about minoritized groups. Despite that, there are no corpora and models specifically designed for this task. In this paper we fill this gap by presenting a new corpus annotated for biographical event detection. The corpus, which includes $20$ Wikipedia biographies, was compared with five existing corpora to train a model for the biographical event detection task. The model was able to detect all mentions of the target-entity in a biography with an F-score of $0.808$ and the entity-related events with an F-score of $0.859$.  Finally, the model was used for performing an analysis of biases about women and non-Western people in Wikipedia biographies.
\end{abstract}

\section{Introduction}
Detecting biographical events from unstructured data is a relevant task to explore and compare bias in representations of individuals. In recent years, the interest in this topic has been favored by studies about social biases on allegedly objective public archives such as Wikipedia. \newcite{sun2021men} developed a resource for investigating gender bias on Wikipedia biographies showing that personal life events tend to be more frequent in female career sections than in those of men. \newcite{lucy2022discovering} developed BERT-based contextualized embeddings for exploring representations of women on Wikipedia and Reddit.  

The detection of biographical events has been addressed with complementary approaches by different research communities. Projects in Digital Humanities have focused mostly on representational aspects, delivering ontologies and knowledge graphs for the collection and study of biographical events 
\cite{tuominen2018bio,fokkens2018biographynet, plum2019large,krieger2014detailed}. 
When it comes to NLP, the focus has been mainly on developing models for the detection and classification of events \cite{rospocher2016building,gottschalk2018eventkg}. 
Few are the works that directly target biographies and focus on identifying biographical events with varied approaches (supervised and unsupervised) across different datasets (e.g., Wikipedia \textit{vs.} newspaper articles), making their comparison impossible~\cite{bamman2014unsupervised,russo2015extracting,menini2017ramble}. Although not directly targeting biographies, some works focused on the identification of entity-related sequences of events~\cite{chambers2008unsupervised} and entity-based storylines~\cite{chambers-jurafsky-2009-unsupervised,minard2015semeval,VOSSEN201660}. 

 
Despite the above mentioned variety of approaches to biographical event detection, 
there are pending and urgent issues to be addressed, which limit a full development of the research area.
In particular, we have identified three critical issues: 
\textit{i)} the lack of 
a benchmark annotated corpus for evaluating biographical event detection; 
\textit{ii)} the lack 
of models specifically designed for detecting and extracting biographical events; 
and finally \textit{iii)} the lack of a systematic study of the potential representation bias of minority groups, non-Western people, and younger generations in  biography repositories publicly available, such as Wikipedia~\cite{d2020data}.


\noindent
{\bf Contributions} 
Our work addresses these issues by presenting a novel benchmark corpus, a BERT-based model for biographical event detection, and an analysis of $48,789$ Wikipedia biographies of writers born since 1808. 
%
%
Our results 
show that existing data sets annotated for event detection may be easily re-used to detect biographical events achieving good results in terms of F-measure. The analysis of the $48,789$ biographies from Wikipedia extends the findings from previous work indicating that representational biases are present in an allegedly objective source such as Wikipedia along intersectional axes~\cite{crenshaw2017intersectionality}, namely ethnicity and gender. 


The rest of the paper is organized as follows. In Section~\ref{sec:dataset}, we present \ds, 
a novel manually annotated corpus of biographical events. Section~\ref{sec:model} presents the  experiments in event detection and coreference resolution of the target entities of a biographies. Section~\ref{sec:intersectionalAnalysis} is devoted to the analysis of the biases in Wikipedia biographies. Conclusions and future work end the paper in
Section~\ref{sec:future_work}.

Code and \ds corpus are available at the following url: \url{https://github.com/marcostranisci/WikiBio/}.

\section{The \ds Corpus} \label{sec:dataset}
\ds is a corpus annotated for biographical event detection, composed of $20$ Wikipedia biographies. 
The corpus includes all the events which are associated with the entity target of the biography.

In this section, we present our annotation scheme, 
discuss the agreement scores and present some cases of disagreement. Lastly, we present the results of our annotation effort, and compare them 
with existing corpora annotated for event detection and coreference resolution.

\subsection{Annotation Tasks} \label{ss:guidelines}
Since the biographical event detection task consists in annotating all events related to the person who is the subject of a biography, annotation guidelines focus on two separate subtasks: (i) the identification of all the mentions of the target entity and the resolution of its coreference chains; and 
(ii) the identification and linking of all the events that involve the target entity.

\paragraph{Entity annotation.} The entity annotation subtask requires the identification of all mentions of a specific Named Entity (NE) \cite{grishman1996message} of type Person, which is the target of the biography and all its coreferences~\cite{deemter2000coreferring} within the Wikipedia biography.
For the modeling of this subtask, we used the GUM corpus \cite{zeldes2017gum}, introducing different guidelines about the following aspects: 
\textit{i)} only the mentions of the entity-target of the biography must be annotated; \textit{ii)} mentions of the target entity must be selected only when they have a role in the event (Example~\ref{ex_01}, where the possessives ``his'' is not annotated); and \textit{iii)} indirect mentions of the target entity must be annotated only if they are related to biographical events (Examples~\ref{ex_they} and~\ref{ex_02}). 


\begin{enumerate}
\setcounter{enumi}{0} 
    \item Kenule Saro-Wiwa was born in Bori [...] \textbf{His} father's hometown was the village of Bane, Ogoniland. \label{ex_01}
    \item \textbf{He} married Wendy Bruce, whom \textbf{he} had known since \textbf{they} were teenagers. \label{ex_they}
    \item In 1985, the Biafran Civil War novel \textbf{Sozaboy} was published. \label{ex_02}
\end{enumerate}

\paragraph{Event Annotation.} Although there is an intuitive understanding of how to identify event descriptions in natural language texts, there is quite a large variability in their realizations~\cite{pustejovsky2003timebank}. \citet{araki-etal-2018-interoperable} point out that some linguistic categories, e.g., nouns, 
fits on 
an event \textit{continuum}. This makes the identification of event mentions a non trivial task.
Our event annotation task mainly relies on TimeML \cite{pustejovsky2003timeml} and RED \cite{o2016richer}, 
where `event' is ``a cover term for situations that happen or occur.'' \cite{pustejovsky2003timeml} 

Events are annotated at a single token level with no restrictions on the parts of speech that realize the event. 
Following~\citet{bonial2016comprehensive}, we introduced a special tag (\texttt{LINK}) for marking a limited set of light and copular verbs, as illustrated in Example~\ref{ex_03}. The adoption of \texttt{LINK} is aimed at increasing the compatibility of the annotated corpus with OntoNotes, the resource with the highest number of annotated events.

\begin{enumerate}
    \setcounter{enumi}{3}
    \item Ken Saro-Wiwa $<$\texttt{LINK}$>$\textbf{was}$<$\texttt{LINK}/$>$ a Nigerian $<$\texttt{EVENT}$>$\textbf{writer}$<$\texttt{EVENT}/$>$ \\
    $<$\texttt{LINK source=`be' target =`writer'} /$>$. \label{ex_03}
\end{enumerate} 
Lastly, to enable automatic reasoning  on biographies, we annotate the contextual modality of events~\cite{o2016richer}. In particular, to account for the  uncertainty/hedged modality, i.e., any lexical item that expresses 
``some degree of uncertainty about the reality of the target event''~\cite{o2016richer}, we have defined three uncertainty values: \texttt{INTENTION}, for marking all the events expressing an intention (like `try' or `attempt'); \texttt{NOT\_HAPPENED}, for marking all events that have not occured; \texttt{EPISTEMIC}, which covers all the other 
types 
of uncertainty (e.g., opinion, conditional). The uncertainty status of the events is annotated by linking the contextual modality marker and the target event, as illustrated in Example~\ref{ex_04}:


\begin{enumerate}
    \setcounter{enumi}{4}
    \item Feeling alienated, he \textbf{decided} to \textbf{quit} college, but was \textbf{stopped} [...] \\ $<$\texttt{CONT\_MOD source ='decided' target = quit' value='INTENTION'} / $>$ \\ 
    $<$\texttt{CONT\_MOD source ='stopped' target = 'quit' value='NOT\_HAPPENED'} / $>$ \label{ex_04}
\end{enumerate} 

\paragraph{Corpus Annotation and IAA.}
\begin{table}
\centering
\begin{tabular}{lcc}
\toprule
\textbf{Annotation Layer} & \textbf{A0 \& A1} & \textbf{A0 \& A2}\\
\midrule
Event & $0.72$ & $0.86$ \\
Entity & $0.65$ & $0.86$ \\
LINK & $0.76$ & $0.64$\\
CONT\_MOD & $0.71$ & $0.64$ \\
\bottomrule
\end{tabular}
\caption{Inter-Annotator Agreement (Cohen's Kappa).}
\label{tab:iaa}
\end{table}

The annotation task was performed by three expert annotators (two men and one woman - 
all 
authors of the paper), near-native speakers of British English, having a long experience in annotating data for the specific task (event and entity detection). 
One annotator (A0) was in charge of preparing the data by discarding all non-relevant sentences to speed-up the annotation process. This resulted in a final set of $1,691$ sentences containing at least one mention of a target entity. The entity and event annotations were conducted as follows: A0 annotated the entire relevant sentences, while a subset of $400$ sentences was annotated by A1 and A2, who respectively labeled $200$ sentences each. We report pair-wise Inter-Annotator Agreement (IAA) using Cohen's kappa in Table~\ref{tab:iaa}. In general, there is a fair agreement across all the annotation layers. At the same time, we observe a peculiar behavior across the annotators: there is a higher agreement between A0 and A2 for the event and entity layers when compared to A0 and A1, but the opposite occurs with the relations layers (LINK and CONT\_MOD).

For the events, the higher disagreement is due to nominal events,  often misinterpreted as not bearing an eventive meaning. For instance, the noun ``trip'' in example~\ref{ex_agree} was not annotated by A1.

\begin{enumerate}
    \setcounter{enumi}{5}
    \item When Ngũgĩ \textbf{returned} to America at the end of {\color{red}\textbf{his}} month {\color{red}\textbf{trip}} [...] \label{ex_agree}
\end{enumerate} 

For the entities, we observed that disagreement is due to two reasons. The first is the consequence of a disagreement in the event annotations. Whenever annotators disagree on the identification of an event, they also disagree on the annotation of the related entity mention, as in the case of the pronoun `his' in example~\ref{ex_agree}. Another reason of disagreement regards indirect mentions. Annotators often disagree on annotation spans, as in ``Biafran Civil War novel Sozaboy was published'' where A1 selected `SozaBoy', while A2 `novel Sozaboy'.
When it comes to LINK, problems are mainly due to the identification of light verbs. Despite the decision of considering only a close set of copular and light verbs to be marked as LINK  (Cfr \newcite{bonial2016comprehensive}), annotators used this label for other verbs, such as `begin' or `hold'. 

\begin{enumerate}
    \setcounter{enumi}{6}
    \item Walker \textbf{began} to take up reading and writing. \label{ex_rel}
\end{enumerate} 

\begin{table*}[h]
\centering
\begin{tabular}{lp{2,5cm}lll}
\toprule
\textbf{Corpus} & \textbf{Size} & \textbf{Text types} & \textbf{Relevant task}\\
\midrule
TimeBank & $7,471$ events & news & Event detection \\
OntoNotes & $159,938$ events, $22,234$ entity mentions & frame-theory & Event \& Entity detection\\
NewsReader & $594$ events & TimeML & Event detection\\
GUM & $9,762$ entity mentions & biographies & Entity detection \\
LitBank & $7,383$ events & literary works & Event Detection\\
\bottomrule

\end{tabular}
\caption{A list of five existing resources that have been employed in the biographical event detection task.}
\label{tab:corpora overview}
\end{table*}


\subsection{\ds: Overview and Comparison with Other Resources} \label{ss:corpus_description}


The \ds corpus is composed of $20$ biographies of African, and African-American writers extracted from Wikipedia for a total amount of $2,720$ sentences. Among them, only $1,691$ sentences include at least one event related to the entity target of the biography. More specifically, there are $3,290$ annotated events, $2,985$ mentions of a target entity, $343$ LINK tags, and $75$ CONT\_MOD links. 

\paragraph{Corpora size and genres}
We compare \ds against five relevant existing corpora that, in principle, could be used to train models for biographical event detection: GUM \cite{zeldes2017gum}, Litbank \cite{sims-etal-2019-literary}, Newsreader \cite{minard2016meantime}, OntoNotes \cite{hovy2006ontonotes}, and TimeBank \cite{pustejovsky2003timebank}. For each corpus, we took into account the number of 
relevant annotations 
and the types of texts. As it can be observed in Table \ref{tab:corpora overview}, corpora vary in size and genres. OntoNotes is the biggest one and includes $159,938$ events, and $22,234$ entity mentions. The smaller is NewsReader, with only $594$ annotated events. TimeBank and LitBank are similar in scope, since they both include about $7.5K$ events, while GUM includes $9,762$ entity mentions.

\paragraph{Text types} With the exception of GUM, which includes $20$ biographies out of $175$ documents, all other corpora contains types of texts other than biographies such as news, literary works, and transcription of TV news. To get a high-level picture of the potential similarities and differences in terms of probability distributions, we calculated the Jensen-Shannon Divergence \cite{menendez1997jensen}. Such metric may be useful for identifying which corpora are most similar to \ds.  The results show that \ds converges more with GUM ($0.43$), OntoNotes ($0.48$) and LitBank ($0.49$) rather than with TimeBank ($0.51$) and Newsreader ($0.54$). Such differences have driven the selection of data for the training set described in Section \ref{ss:event_detection}.


\paragraph{Annotations of entities, events, and coreference}

The distribution of the target entity within biographies in the \ds corpus has been compared with two annotated corpora for coreference resolution and named entity recognition: OntoNotes \cite{hovy2006ontonotes} and GUM \cite{zeldes2017gum}. Since such corpora were developed for identifying the coreferences of all NEs in a document, we modified annotations to  keep only the most  
frequent NEs of type `person' in each document. The rationale was making these resources comparable with \ds,  which  includes only the coreferences to a single entity, namely the subject of each biography. After doing that, we computed the ratio between the number of tokens that mention the target entity and the total number of tokens, and the ratio between the number of sentences where the target entity is mentioned against the total number of sentences. While this operation did not impact on GUM, in which $174$ out of $175$ documents contain mentions of people, it had an important impact on OntoNotes, in which $1,094$ documents ($40\%$) do not mention entities of the type Person. \\
Tokens mentioning the target entity are $5\%$ on OntoNotes, $8.7\%$ on GUM and $4\%$ on \ds.  Such differences can be explained by the average length of documents in these corpora, which is of $388$ tokens in OntoNotes, $978$ in GUM, and $3,754$ in \ds. As a matter of fact, if the percentage of sentences mentioning the target-entity is considered instead of the total number of tokens, \ds shows an higher ratio ($61.7\%$) of sentences mentioning the target entity, than OntoNotes ($20.8\%$) and GUM ($42.6\%$).

\begin{table*}[h]
\centering
    \resizebox{.75\textwidth}{!}{
\begin{tabular}{l|llllll}
\toprule
& \textbf{\ds} & \textbf{GUM} & \textbf{Litbank} & \textbf{Newsreader} & \textbf{OntoNotes} & \textbf{Timebank} \\
\midrule
\textbf{\ds} & $0.00$ & $0.43$ & $0.49$ & $0.54$ & $0.48$ & $0.51$ \\
\textbf{GUM} & $0.43$ & $0.0$ & $0.49$ & $0.54$ & $0.39$ & $0.49$ \\
\textbf{Litbank} & $0.49$ & $0.49$ & $0.00$ & $0.55$ & $0.42$ & $0.51$ \\
\textbf{Newsreader} & $0.54$ & $0.55$ & $0.54$ & $0.00$ & $0.48$ & $0.45$ \\
\textbf{OntoNotes} & $0.48$ & $0.39$ & $0.42$ & $0.48$ & $0.00$ & $0.40$ \\
\textbf{TimeBank} & $0.51$ & $0.49$ & $0.51$ & $0.45$ & $0.40$ & $0.00$ \\
\bottomrule
\end{tabular}
}
\caption{The similarity between corpora for event annotation computed with the Jensen-Shannon Divergence.}
\label{tab:jensen-shannon}
\end{table*}

The three most frequently occurring lemmas in the \ds corpus seem to be strongly related to the considered domain: `write' represents $3.2\%$ of the total, `publish'  $2.9\%$, and `work' $1.8\%$. `Return' ($1.3\%$) appears to have a more general scope, since it highlights a movement of the target entity from a place to another. The comparison with other corpora annotated for event detection shows differences concerning the most frequent events. The top three in OntoNotes \cite{bonial2010propbank} are three light verbs: `be', `have', and `do'. This may be intrinsically linked to its annotation scheme which considers all verbs as candidates for being events, including semantically empty ones (Section \ref{ss:guidelines}). NewsReader \cite{minard2016meantime} and TimeBank \cite{pustejovsky2003timebank} include two verbs expressing reporting actions among the top five, thus revealing that they are corpora of annotated news. Litbank \cite{sims-etal-2019-literary}, which is a corpus of $100$ annotated novels, includes in its top-ranked events two visual perception verbs and two verbs of movement, which may reveal the centrality of characters in this documents. The event `say' is top-ranked in all the five corpora.


\section{Detecting Biographical Events} \label{sec:model}

In this section we describe a series of experiments for the detection of biographical events. Experiments involve the use of the existing annotated corpora for two tasks: entity mentions detection (Section \ref{ss:entity_detection}) and event detection (Section \ref{ss:event_detection}). In both cases we used a 66 million parameters DistilBert model \cite{sanh2019distilbert}.  In this setting the \ds corpus is both used as part of the training set and as a benchmark for testing how well existing annotated corpora may be used for the task. For such experiments a NVIDIA RTX 3030 ti was used. The average length of each fine-tuning session was $40$ minutes.

\begin{table*}[h]
\centering
    \resizebox{.85\textwidth}{!}{
\begin{tabular}{llll}
\toprule
\textbf{Training | Dev | Test (30 EPOCHS)} & \textbf{F-Score\_train} & \textbf{F-Score\_dev} & \textbf{F-Score\_test}\\
\midrule
Gum | \ds | \ds & $0.820$ & $0.728$ & $0.752$ \\
Gum+\ds | \ds | \ds & $0.819$ & $0.728$ & $0.753$ \\
Onto | \ds | \ds & $0.896$ & $0.782$ &\textbf{ 0.808} \\
Onto+\ds | \ds | \ds & $0.846$ & $0.774$ & $0.800$ \\
Misc | \ds | \ds & $0.824$ & $0.766$ & $0.792$ \\
Misc+\ds | \ds | \ds & $0.828$ & $0.764$ & $0.789$ \\
\bottomrule
\end{tabular}
}
\caption{Results of entity detection experiments.}
\label{tab:experimental_mapping_entities}
\end{table*}

\subsection{Entity Detection} \label{ss:entity_detection}
For this task we adapted the annotations in OntoNotes \cite{hovy2006ontonotes} and GUM \cite{zeldes2017gum} keeping only mentions of the most frequent entities of type `person'. 
As a result we obtained $870$ documents from OntoNotes, $174$ from GUM.

The \ds corpus was split into three subsets: five documents for the development, $10$ for the test, and five for the training.  Given the imbalance between the existing resources and \ds, we always trained the model with a fixed number of $100$ documents, in order to reduce the overfitting of the model over the other datasets.

Experiments consist in training a DistilBert model for identifying all the tokens mentioning the target entity of a given model and were performed on six different training sets. Since the focus of our work is to develop a model for detecting biographical events, \ds was used as development set for better monitoring its degree of compatibility with existing corpora.
Following the approach by~\newcite{joshi2020spanbert}, we split each document into sequences of $128$ tokens, and for each document we created one batch of variable length containing all the sequences. 
Table \ref{tab:experimental_mapping_entities} shows the results of these experiments. As it can be observed, including the \ds corpus in the training set did not result in an increase of the performance of the model. This may be due to the low number of \ds documents in the training.
The highest performance was obtained in two experiments: one using
a training set only composed of documents from OntoNotes, which obtained a F-score of $0.808$, and 
one with 
a miscellaneous of $50$ OntoNotes and $50$ GUM documents, that obtained $0.792$. To understand if the difference between the two experiments is significant, we performed a One-Way ANOVA test over the train, development, and test F-scores obtained in both experiments. The test returned a p-value of $0.44$, which confirms a significant difference between the two results 
%

\subsection{Event Detection} \label{ss:event_detection}
Event Detection experiments were guided by the comparison between \ds and the resources for event detection described in Section \ref{ss:corpus_description}. Since OntoNotes was annotated according to the PropBank guidelines \cite{bonial2010propbank}, which only consider verbs as candidates for such annotation, we partly modified its annotations before running the experiments. We first adapted the OntoNotes semantic annotation by replacing light and copular verbs \cite{bonial2016comprehensive} with nominal \cite{meyers2004nombank} and adjectival events. Then we ran a battery of experiments by fine-tuning a DistilBert-based model using each dataset for training, and a series of miscellaneous of the most similar corpora to \ds according to the Jensen-Shannon Divergence metric (Table~\ref{tab:jensen-shannon}).
Since we were concerned with both assessing the effectiveness of \ds for training purposes and  testing   how far biographic events can be extracted, we designed our training and testing data as follows. \ds was employed in different learning phases: in devising the training set (i.e., existing resources were employed either alone or mixed with \ds); additionally, the development set was always built by starting from \ds sentences. Finally, we always tested on \ds data.




As for the entity-detection experiments, the $1,691$ sentences containing events annotated in the \ds corpus were split into three sets of equal size that were used for training ($564$), development ($563$), and testing ($564$). Given the disproportion between  OntoNotes and other corpora, we sampled a number of sentences for training which did not exceeded $5,073$, namely three times the number of sentences annotated in our corpus. Such length was fixed also for miscellaneous training sets. 

Experiments were organized in two sessions. In the first session we fine-tuned a DistilBert model for five epochs, using as training set the five corpora presented in Section \ref{ss:corpus_description} individually as well as three combinations of them: \textit{i)} misc\_01, a miscellaneous of sentences extracted on equal size from all corpora; \textit{ii)} misc\_02, in which sentences from NewsReader, the most different corpus with \ds (Table \ref{tab:jensen-shannon}), were removed; \textit{iii)} misc\_03, a combination of sentences from OntoNotes and Litbank, namely the two most similar corpora with \ds. The model was fine-tuned on these training sets both with and without a subset of the \ds corpus for a total of $16$ different training sets. In addition, we also fine-tuned and tested \ds alone. We then continued the fine-tuning only for the models which obtained the best F-scores. 

Observing Table \ref{tab:full_report}, it emerges that, differently from entity-detection experiments, including a subset of \ds in the training set, even if in a small percentage, always improves the results of the classifier. This especially happens for Litbank ($+0.191$ F-Score), and TimeBank ($+0.031$ F-Score).

When looking at results of finetuning for single corpora, it emerges that the model trained on the modified version of OntoNotes and TimeBank obtains the best scores. Such results are interesting for two reasons. They confirm the intuition that OntoNotes annotations may be easily modified to account for nominal and adjectival events. They also confirm the high compatibility of \ds and TimeBank guidelines (Sect. \ref{ss:guidelines}). Even if the latter is more divergent from \ds than other corpora, it seems to be compatible with it. As expected for its limited size and high divergence with \ds, the training set based on NewReader sentences obtains the worst results, with an F-Score below $0.5$.

Results of miscellaneus training sets are interesting as well: they generally result in models with better performance, and they seem to work better on the basis of their divergence with \ds. Trained on misc\_01, a combination of all corpora, the model scores $0.827$, which is below the result obtained with the modified version of OntoNotes. If Newsreader is removed, the model obtains $0.831$, and $0.832$ if also TimeBank is removed. It is also worth mentioning the delta between the F-score on the training and the test sets, which is $-0.054$ for misc\_01, $-0.029$ for misc\_02, and $-0.013$ for misc\_03.


After the first fine-tuning step, we performed a One-Way ANOVA for testing the significance of differences between experiments. Analyzed in such a way, the four best-ranked models never showed a p-value below $0.5$, which means that there are no significant differences between them. Thereby, we kept them for the second fine-tuning step that consists on training the model for $15$ epochs on these datasets. Absolute results (Table \ref{tab:full_report}) show that the model trained on Timebank obtained the best F-Score. However, as for the entity detection experiments, we considered the deltas between the training and test F-scores to select the best model for our analysis. All models acquired by employing a miscellaneous training set obtained a lower delta between training and test, and scored a similar F-Score.   

\begin{table*}[ht]
\centering
    \resizebox{.85\textwidth}{!}{
\begin{tabular}{l|lll}

\textbf{Training | Dev | Test (5 EPOCHS)} & \textbf{F-Score\_train} & \textbf{F-Score\_dev} & \textbf{F-Score\_test}\\
\hline
\hline
\ds | \ds | \ds & $0.479$ & $0.479$ & $0.479$ \\
Litbank | \ds | \ds  & $0.847$ & $0.640$ & $0.622$ \\
Litbank + \ds | \ds | \ds & $0.835$ & $0.814$ & $0.813$ \\
Misc\_01 | \ds | \ds & $0.885$ & $0.863$ & $0.801$ \\
Misc\_01 + \ds | \ds | \ds  & $0.871$ & $0.831$ & $0.827$ \\
Misc\_02 | \ds | \ds & $0.866$ & $0.816$ & $0.819$ \\
Misc\_02 + \ds | \ds | \ds  & $0.861$ & $0.837$ & \textbf{0.832} \\
Misc\_03 | \ds | \ds & $0.850$ & $0.811$ & $0.817$ \\
Misc\_03 + \ds | \ds | \ds  & $0.844$ & $0.839$ & $0.831$ \\
Onto | \ds | \ds & $0.950$ & $0.800$ & $0.790$ \\
Onto + \ds | \ds | \ds & $0.936$ & $0.873$ & $0.809$ \\
Onto\_mod | \ds | \ds & $0.997$ & $0.823$ & $0.814$ \\
Onto\_mod + \ds | \ds | \ds & $0.888$ & $0.869$ & $0.829$ \\
Timebank | \ds | \ds  & $0.89$ & $0.801$ & $0.790$ \\
Timebank + \ds |\ds | \ds & $0.865$ & $0.856$ & $0.821$ \\
NewsReader | \ds | \ds & $0.453$ & $0.479$ & $0.479$ \\
NewsReader + \ds | \ds | \ds & $0.467$ & $0.479$ & $0.479$ \\

\hline
\hline
\textbf{Training | Dev | Test (15 EPOCHS)} & \textbf{F-Score\_train} & \textbf{F-Score\_dev} & \textbf{F-Score\_test}\\
Misc\_01 + \ds | \ds | \ds  & $0.890$ & $0.852$ & $0.853$ \\
Misc\_02 + \ds | \ds | \ds & $0.900$ & $0.855$ & $0.856$ \\
Misc\_03 + \ds | \ds | \ds & $0.896$ & $0.859$ & $0.855$ \\
Timebank + \ds | \ds | \ds & $0.919$ & $0.850$ & \textbf{0.859} \\
\end{tabular}
}
\caption{Results of event detection experiments: complete table}
\label{tab:full_report}
\end{table*}

\section{An Intersectional Analysis of Wikipedia Biographies}\label{sec:intersectionalAnalysis}

In this section we provide an analysis of writers' biographies on Wikipedia adopting intersectionality as a theoretical framework and the model described in Section \ref{sec:model} as a tool for detecting biographical events.
%

The concept of intersectionality \cite{crenshaw2017intersectionality} has been developed in the context of gender and black studies to account inequalities that cannot be explained without a joint analysis of socio-demographic factors. For instance, African American women workers suffer higher discrimination than their male counterpart, as \newcite{crenshaw1989demarginalizing} observed in her seminal work. 
Therefore, the injection of different socio-demographic features for the analysis of discriminations may unfold hidden forms of inequities about certain segments of population. We adopt this framework to analyse how the representations of non-Western women writers on Wikipedia differs from those of Western Women, Transnational Men, and Western Men.

For this analysis, we gathered $48,486$ Wikipedia biographies of writers born since $1808$. 
We define as Transnational all the writers born outside Western countries and people who belong to ethnic minorities~\cite{boter2020unhinging,stranisci2022urw}. Western men's biographies are $28,036$, Western women's $12,413$, Transnational men's $5,471$, and Transnational women's $2,470$. 
Information about occupation, gender, year of birth, ethnic group, and country of birth was obtained from Wikidata \cite{vrandevcic2014wikidata}, which has been used for filtering and classifying biographies.

For each biography, we first identified all the mentions of the corresponding target entity (Section \ref{ss:entity_detection}). We then removed the sentences that do not contain a mention of the entity. This reduced the number of sentences to be annotated for event detection from $1,486,320$ to $1,163,475$ ($-21.8\%$). As a final step, we annotated events (Section \ref{ss:event_detection}) in the filtered sentences. 

Table \ref{tab:events_gender_ethnicity} shows the distribution of biographical events about men, women, Western, and Transnational people. The vast majority of events are about Western men ($62.2\%$), while at the opposite side of the spectrum there are Transnational women writers, whose representation is below $5\%$. Ethnicity is a cause of underrepresentation more than gender: events about Transnational men are only $11.2\%$ of the total, while those about Western women $21.4\%$. The average number of events per-author shows a richness in the description of Transnational Women ($50.92$ events) against Western ones ($43.73$ events).

The analysis of event types presents a similar distribution. $27,885$ event types -- intended as the number of unique tokens that occur in each distribution -- are detected in Western men's biographies ($44.9$ \textit{per} biography), while only $9,254$ in Transnational women's biographies ($40.4$ \textit{per} biography). 
However, the overlap of event types between these two categories is very large ($92.6\%$)
The same comparison, conducted on the other groups, reveals a higher number of group-specific event types: $87.8\%$ of event types about Transnational Men are shared with Western Men, and the rate is lower for Western Women ($84.1\%$). 

\begin{table}[b]
\centering
\begin{tabular}{llll}
\toprule
\textbf{Group} & \textbf{Events} & \textbf{Avg} & \textbf{Types}\\
\midrule
\textbf{Western M} & $1,57M$ & $56.08$ &  $27,885$ \\
\textbf{Transnational M} & $285K$ & $52.10$ &  $14,057$ \\
\textbf{Western W} & $542K$ & $43.73$ & $17,324$ \\
\textbf{Transnational W} & $125K$ & $50.92$ &  $9,254$  \\
\bottomrule
\end{tabular}
\caption{The number of events and event types broken down by gender and ethnicity of writers.}
\label{tab:events_gender_ethnicity}
\end{table}

A comparative analysis of most distinctive events per category of people provides additional insight about the representation of women and Transnational writers in Wikipedia biographies. In order to do so, we first computed the average frequency of each event in all biographies of the four groups of writers in Table \ref{tab:events_gender_ethnicity}. We then compared these distributions with the Shifterator library \cite{gallagher2021generalized}, which allows computing and plotting pairwise comparisons between different distribution of texts with different metrics. Coherently with the analysis performed in previous sections, we chose the Jensen-Shannon Divergence metric, and analyzed the distribution of events about Transnational Women against Transnational Men, Western Men, and Western Women. Table \ref{tab:wst_und} shows the most diverging events between Transnational and Western writers, while Table \ref{tab:comparisons} shows the $20$ events about Transnational women that diverge most with other distributions: Transnational men, Western men, and Western women. Events are ordered on the basis of how much they are specific to the distribution of Transnational women. In Appendix \ref{app:comparisons} graphs with comparisons between distributions can be consulted.

A first insight from a general overview of distinctive events about Transnational Women writers 
is that they seem to 
never
die. Events like `death' or `died' are never distinctive for them but always for the group against which they are compared. This may be explained by the average year of birth of Transnational Women writers with a biography on Wikipedia, which is $1951$, while for Western men is $1936$, $1943$ for Transnational men, and $1944$ for Western woman. 
%
%

The analysis of the most salient biographical events between Transnational women and Transnational men shows how intersectionality helps to identify gender biases. 
When Transnationals are considered as a single group (Table \ref{tab:wst_und}) against the Western counterparts, the majority of the biographical events are related to career (award, conferred) or to social commitment (activist, migrated, exile). When the comparison is made within the Transnational group (Table \ref{tab:comparisons}), the gender bias demonstrated by \newcite{sun2021men} and \newcite{bamman2014unsupervised} clearly emerges. In fact, `married', `marriage', and `divorce' are associated to Transnational women. In addition, there is a lack of career-related events about them, while this is not the case for men (actor, chairman, politician). The comparison between Transnational women and Western men still shows a gender bias, but less prominent. Among the most salient events, only  `mother’ highlights a potential bias, while events on Transnational women career (`win', `won', `award', `selected'), education (`degree', `education', `schooling') and social commitment (`activist’) are present. \\
\indent Finally, the comparison between Transnational and Western women offers three additional insights. First, the only event about private life which is salient for one of the two groups is `married'. This indicates that private life events of women - in general -  are always presented in relation to their conjugal status. Second, 
careers and social commitments are particularly present for Transnational women. 
Finally, 
the framing of the concept ``relocation'' is expressed using different event triggers: 
the more neutral `move' is used for Western women, while the more marked, negatively connotated term `migrate' is associated with Transnational women.\\ 
\indent Summarizing, Transnational Women are underrepresented on Wikipedia with respect to other groups, both in terms of number of biographies and events. 
The analysis of their most distinctive biographical events shows that the already-known tendency of mentioning private life events about women in Wikipedia biographies \cite{sun2021men,bamman2014unsupervised} can be refined when coupled to ethnic origins. 
Indeed, the extent of the presence of gender biases is more salient when comparing the biographical entries within the same broad ``ethnic'' group, while is becomes obfuscated across groups, making other bias (i.e., racial) more prominent. 


\section{Conclusion and Future Work}\label{sec:future_work}
In this paper we presented a novel set of computational resources for deepening the analysis of biographical events and improving their automatic detection. We found that existing annotated corpora may be successfully reused to train models that obtain good performances. The model for entity detection, trained on OntoNotes, obtained a F-score of $0.808$, while the model for event detection, trained on TimeBank and Wikibio, scored $0.859$. We have applied these newly developed resources to 
perform an analysis of biases in Transnational women writers on Wikipedia adopting intersectionality as a framework to interpret our results. In particular, we have identified that the representation of women and non-Western people on Wikipedia is problematic and biased. Using different axes of analysis - as suggested by intersectionality - it becomes easier to better identify these biases. For instance, gender bias against Transnational women are more marked when comparing their biographies against those of  Transnational men rather than Western ones. On the other hand, potential racial biases emerge when comparing Transnational women to Western women. Using an intersectional framework would benefit the understanding and countering of biases 
of women and non-Western people on Wikipedia. 

Future work will improve the model for biographical event detection, and to extend the analysis on a wider set of biographical entries from different sources.

\begin{table}[!bh]
\centering
{
\begin{tabular}{|p{3.6cm}|p{3.6cm}|}
\toprule
\textbf{Transnational} &\textbf{Western} \\
poet, education, schooling, award, degree, completed, awarded, activist, obtained, professor, started, translated, conferred, migrated, exile, recipient, born, novelist, writer, lyricist & wrote, appeared, sold, illustrated, described, married, starred, met, told,illustrator, enlisted \\

\hline
\end{tabular}
}
\caption{Comparison of biographical events between Transnational and Western writers.}
\label{tab:wst_und}
\end{table}

\begin{table}[!th]
\centering
{
\begin{tabular}{|p{3.6cm}|p{3.6cm}|}
\toprule
\textbf{Transnational Women} & \textbf{Transnational Men}\\
defeated, daughter, actress, married, lost, appeared, marriage, deafeating, won, began, activist, loosing, divorced, raised, attended, win, featured, seeded, mother, grew &
actor, son, chairman, lyricist, served, politician, critic, father, joined, death, accused, known, poet, scholar, elected, imprisoned, president, established, exile\\
\midrule
\textbf{Transnational Women} & \textbf{Western Men}\\
activist, degree, won, actress, received, born, daughter, award, education, defeated, recipient, defeating, win, selected, mother, writer, schooling, completed, poet, lost 
& wrote, enlisted, service, actor, claimed, father, assigned, drafted, directed, developed, death\\

\midrule
\textbf{Transnational Women} & \textbf{Western Women}\\

defeated, defeating, lost, activist, education, loosing, schooling, degree, poet, completed, win, seeded, injury, award, match, reach, migrated, participated, professor, loss 
& wrote, appeared, married, author, published, starred, death, lives, moved, died, sold, illustrator, illustrated, nominated, reviewer, write, lived, developed, spent\\

\hline
\end{tabular}
}
\caption{Comparison of biographical events about Transnational Women \textit{vs.} Transnational Men, Western Men, and Western Women. The tokens in the Table cells were obtained by maximizing the JSD divergence. 
We used the Shifterator software library (see Appendix~\ref{app:comparisons} for details).}
\label{tab:comparisons}
\end{table}


\section*{Limitations and Ethical Issues}
This work presents some limitations that will be addressed in future work. In particular, \textit{i)} even if the model for biographical event detection obtained good results, more sophisticated approaches may be devised to increase its effectiveness (e.g., best performing LMs, multi-task settings); 
\textit{ii)} the intersectional analysis was performed on a specific sample of people, and thus limited to writers. Taking into account people with other occupations may lead to different results; 
finally, \textit{iii)} only Wikipedia biographies were considered: biographies from other sources may differ in style and thus pose novel challenges to the biographical event detection task.

The research involved the collection of documents from Wikipedia, which are released under the Creative Commons Attribution-ShareAlike 3.0 license. The annotation of the experiment was not crowdsourced. All the three annotators are member of the research team who carried out the research as well as authors of the present paper. They are all affiliated with the University of Turin with whom they have a contract regulated by the Italian laws. Their annotation activity is part of their effort related to the development of the present work, which was economically recognized within their contracts with the University of Turin. A data statement for the research can be accessed at the following url: \url{https://github.com/marcostranisci/WikiBio/blob/master/README.md}
\bibliography{anthology,custom}
\bibliographystyle{acl_natbib}

\appendix

\section{Comparison Between Transnational Women and Men through the JS Divergence Metric} \label{app:comparisons}
In this Section you can observe a comparative analysis of the divergence between events about Transnational women against Transnational men (Figure \ref{fig:gender1}), Western men (Figure 2), and Western women (Figure 3). All divergences were computed and plotted with Shifterator \cite{gallagher2021generalized}.
\label{sec:appendix}
\begin{figure*} \label{fig:gender1}
    \includegraphics[width=\textwidth]{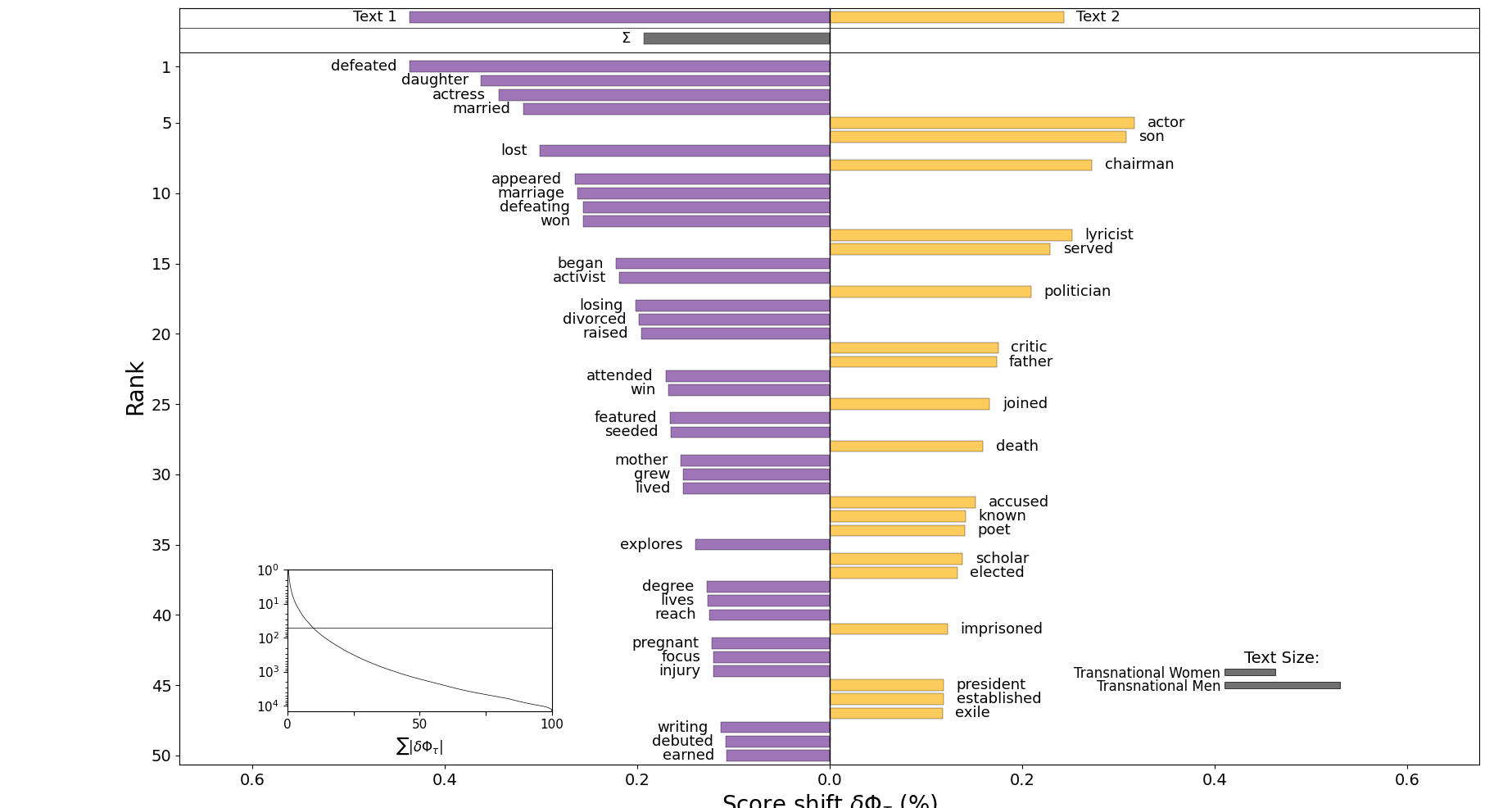}
    \caption{The comparison of events between Transnational Women biographies and Transnational Men biographies.}
    \label{fig:gender1}
\end{figure*}

\begin{figure*} \label{fig:gender2}
    \includegraphics[width=\textwidth]{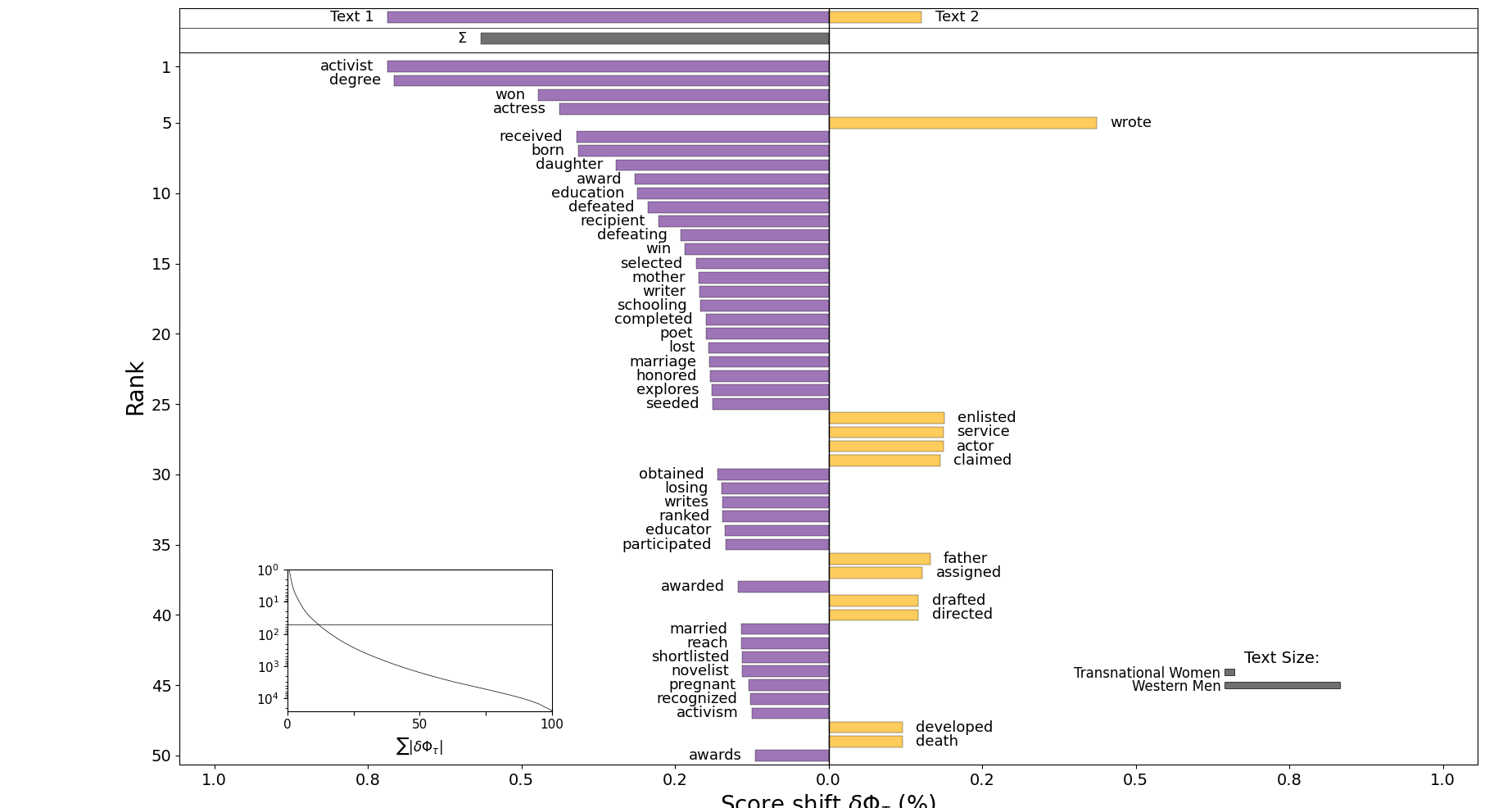}
    \caption{The comparison of events between Transnational Women biographies and Western men biographies.}
    
\end{figure*}

\begin{figure*} \label{fig:gender3}
    \includegraphics[width=\textwidth]{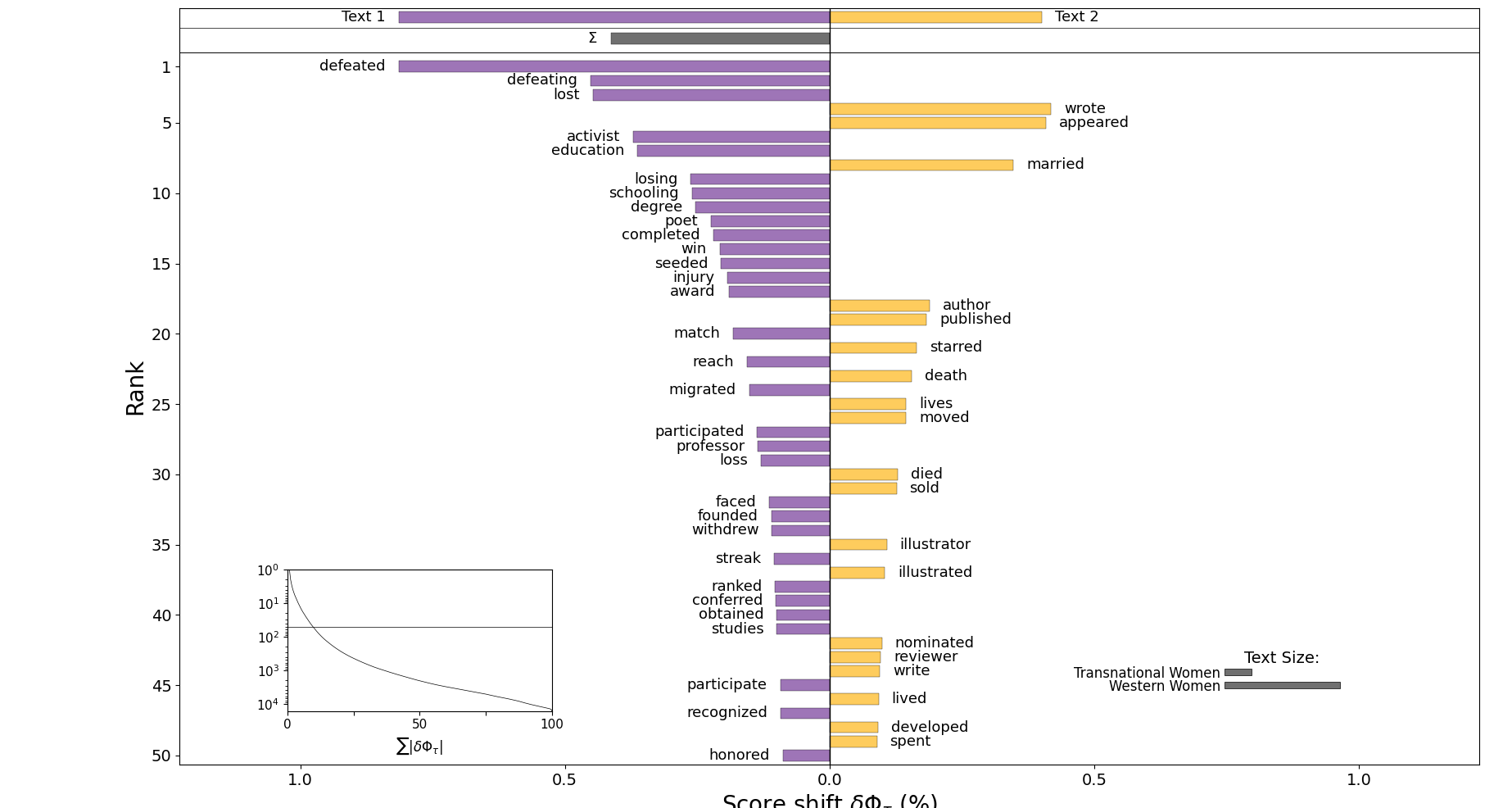}
    \caption{The comparison of events between Transnational Women biographies and Transnational Women.}
    
\end{figure*}

\end{document}